# Corner Cases: Headland Coverage Path Planning for Autonomous Driving in Arable Farming

**Riikka Soitinaho** * **Timo Oksanen** *

* *Technical University of Munich, Germany; Chair of Agrimechatronics; Munich Institute of Robotics and Machine Intelligence (MIRMI), (e-mail: firstname.surname@tum.de)*

**Abstract:** This paper presents a new method for headland coverage path planning for arable fields. Several earlier approaches suggest covering the headland with nested polygons and smooth turns, however, covering the field corners entirely requires manoeuvres with reversing. In the new method, the polygon corners are modified to allow a reversing turn. A comparison to two other methods considering gap, overlap, and crossing the field boundary shows an improvement in the coverage result especially in field corners of around 90 degrees, and 240 degrees and above. Applicability of the new method is shown with several examples of real polygonal field maps.

## 1. INTRODUCTION

Coverage path planning (CPP) is an important task in automating arable farming for operations such as harrowing and seeding. Possibly the simplest solution is to travel back-and-forth across the field, along parallel tracks exactly one implement width apart. Due to kinematic and dynamic contraints the farm vehicles cannot turn in place, but need extra space to manoeuvre between the tracks. Headland is an area along the field perimiter, where the vehicles can turn from one track to another. To maximize the yield, also the headland needs to be covered.

The headland coverage problem is often addressed together with the mainland coverage problem. Höffmann et al. (2024) suggest that the headland tracks are most often planned with nested polygons (e.g. Hameed et al. (2010), Nilsson and Zhou (2020)), and the polygon corners are modified with another curve that has continuous curvature (e.g. Höffmann et al. (2022)). These methods result in rounding the corners of the field, leaving a part of the corners uncovered. Fully covering the headland would require manouvres with reversing (Nilsson and Zhou (2020)). Pour Arab et al. (2023) find Edwards et al. (2017) and Jeon et al. (2021) to be the only works that consider reverse manoeuvres to cover the field corners. More recently, Mier et al. (2025) propose a headland coverage method including an X-type turn that includes reversing. While the method considers reversing, it cannot fully cover the outermost headland track in the field corner due to the vehicle dimensions, and it requires an offset from the field boundary that introduces a gap along the sides of the field.

While there is already a lack of literature focusing on headland coverage, especially the field corners tend to be overlooked. Based on the findings of Nilsson and Zhou (2020), Pour Arab et al. (2023), Höffmann et al. (2024), Mier et al. (2025), and our own investigation, we conclude that there is still a lack of a complete description of a headland coverage path planning method in the previous literature, especially one that considers fully covering the headland corners.

The method of Mier et al. (2025) consolidates coverage planning with planning the fieldwork pattern and the turn paths. Another approach is to decouple the three planning tasks, a strategy often applied in coverage path planning for the mainland. Solving the coverage of a field involves planning the coverage tracks and their orientation (e.g. Zhou et al. (2014)), which can then be traversed in alternative sequences to improve the efficiency of the fieldwork (Bochtis and Vougioukas (2008)). Turn path planning focuses on solving a path from point A to point B, such as between consecutive tracks (e.g. Backman et al. (2015)).

Following this strategy, the method proposed in this paper focuses on coverage paths that improve coverage in the field corners. The key contribution is a new headland coverage path planning method that constructs the headland tracks in a way that allows the implement to reach closer to the field corners than previous methods. In the field corners, which are determined based on where the vehicle has to turn, the vertices of the nested polygons are modified so that instead of polygons, the headland tracks become polylines that reach the field boundary. The idea has been touched upon by Soitinaho et al. (2024), however, a complete method was not given. The proposed method is compared to an instance of the method with rounded corners, and an adaptation of the X-type turns proposed by Mier et al. (2025). The results show that the proposed method offers an improvement concerning gap, overlap, and crossing in a range of field corner angles.

* This research was funded by the Deutsche Forschungsgemeinschaft (DFG, German Research Foundation) - 528103308.

## 2. METHOD AND IMPLEMENTATION

### 2.1 Definitions and scope

The field is assumed to be represented as a simple polygon, a closed polygonal chain. The term polyline is used for an open polygonal chain. Each vertex of a polygon is associated with a vertex angle, the angle formed by two line segments that meet at a vertex. The term corner is used to refer to the general region of a corner in the field polygon. While a vertex is a single point, a corner can consist of multiple vertices, and we assume a field corner has one critical vertex at most.

The final headland coverage paths are referred to as headland tracks. As an intermediate step in the algorithm, the outlines of the headland tracks are calculated from the field boundary with polygon offsetting. To distinguish between the polygon outlines of the headland tracks and the final headland tracks, the polygon outlines are referred to as headland track polygons.

A non-holonomic vehicle cannot perfectly follow polygonal headland tracks. However, when the change of direction between two consecutive line segments is small, the vehicle can follow the headland track with a deviation that is negligible in practice. Otherwise, the vehicle needs to make a turn. Whether the change of direction is "small" is determined based on so-called criticality of a vertex.

The proposed method considers agricultural vehicle-implement combinations where the implement is rigidly mounted to the vehicle with a hitch. The implement can be mounted either in the front or rear of the vehicle. There is no offset between the centre line of the implement and the centre line of the vehicle. The dimensions of the vehicle are modelled by its bounding box. The implement is modelled as a line through its centre point along the direction of its operating width, and an area is considered covered once this line has swept over it.

### 2.2 Number of headland track polygons

The algorithm begins by determining the width of the headland tracks and the number of headland track polygons. While the headland track width is equal to the operating width of the implement ($w_{i,o}$), the number of headland track polygons ($n_t$) depends on the required headland width. The vehicle-implement combination must have enough space to turn between the mainland tracks, and the proposed headland CPP method needs enough space to turn in the headland corners. The number of headland track polygons times the headland track width must cover the required headland width.

This paper focuses on the headland coverage path planning algorithm, and therefore a simple estimate of a predefined number of headland tracks is used in the examples.

### 2.3 Generating the headland track polygons

In the next step of the algorithm, the outlines of the headland tracks are generated with a polygon offsetting algorithm. Symbol $\delta$ denotes the offset distance. When $\delta < 0$, the $\delta$-offset polygon is referred to as an inner offset polygon, meaning that the offset polygon is inside the original polygon. A true offset polygon means that the boundary of the offset polygon is exactly at distance $\delta$ from the boundary of the original polygon. This means that circular arcs appear as the offsets of concave vertices in inner offset polygons. In a linearised offset polygon, the circular arcs are replaced by concave vertices.

Altogether $n_{t+1}$ linearised inner offset polygons are calculated, where the first $n_t$ offset polygons are the outlines of the headland tracks, and the +1 is to calculate the headland-mainland boundary. The offset distance $\delta$ depends on $w_{i,o}$. The outermost headland track polygon is at a distance of $w_{i,o}/2$ from the field boundary. The following headland track polygons are offset from the previous headland track by a distance of $-w_{i,o}$. Finally, to calculate the headland-mainland boundary, the innermost headland track polygon is offset by $(-w_{i,o})/2$.

It is possible that vertices disappear in the offset polygons, i.e. the original field boundary polygon has more vertices than the headland track polygons. The polygon offsetting is assumed to not result in complex i.e. self-intersecting polygons, or to split the polygons into multiple parts.

### 2.4 Critical vertices of the headland track polygons

In the next step of the algorithm, all the critical vertices of all the headland track polygons are determined based on the critical angle threshold ($\gamma_t$), which essentially describes the acceptable heading change between consecutive path segments. The critical angle threshold should be decided based on the maneuvrability of the vehicle-implement combination during operation.

When a wide range of vertex angles is considered critical, there are more critical vertices in the polygon, and therefore it is more likely that nearby vertices are critical. Several nearby critical vertices make it more challenging to modify the polygons in the following steps of the algorithm. It is assumed that no critical vertices have appeared or disappeared in the polygon offsetting, i.e. the field boundary, headland tracks, and the headland-mainland boundary all have the corresponding critical vertices.

At every critical vertex the vehicle has to make a turn, and therefore, the polygon needs to be cut and modified to accommodate for the turn. Turn path planning is considered to be a separate problem that is not addressed in this paper. A suitable turn path planning method has been shown e.g. by Väyrynen (2019). This turn path algorithm is able to plan paths for the vehicle to reverse to the edge of the field boundary (see Fig. 1).

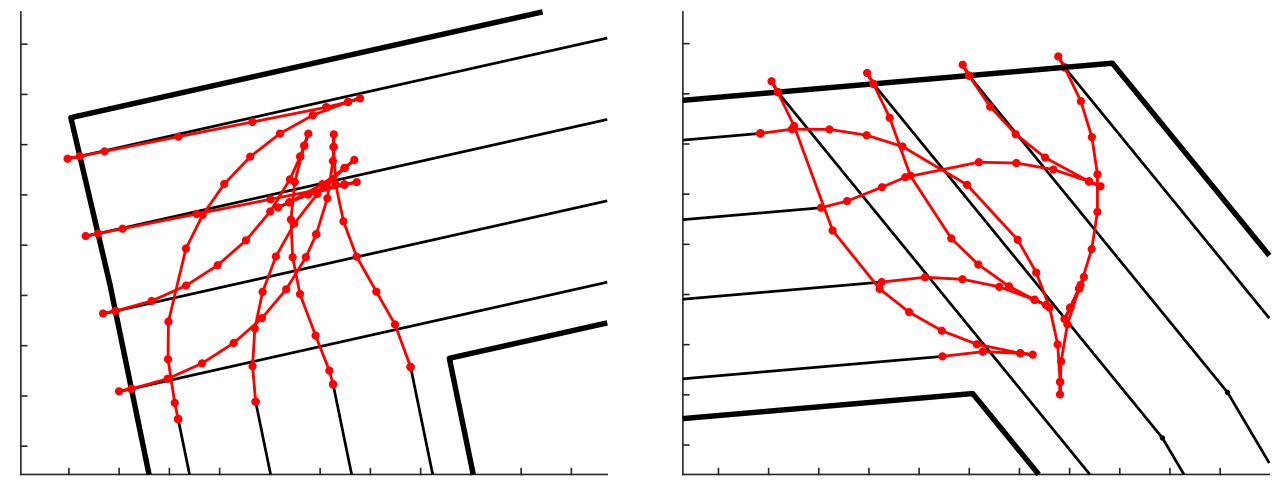

Fig. 1. Headland tracks (black), with turn paths (red) calculated with a method similar to Väyrynen (2019).

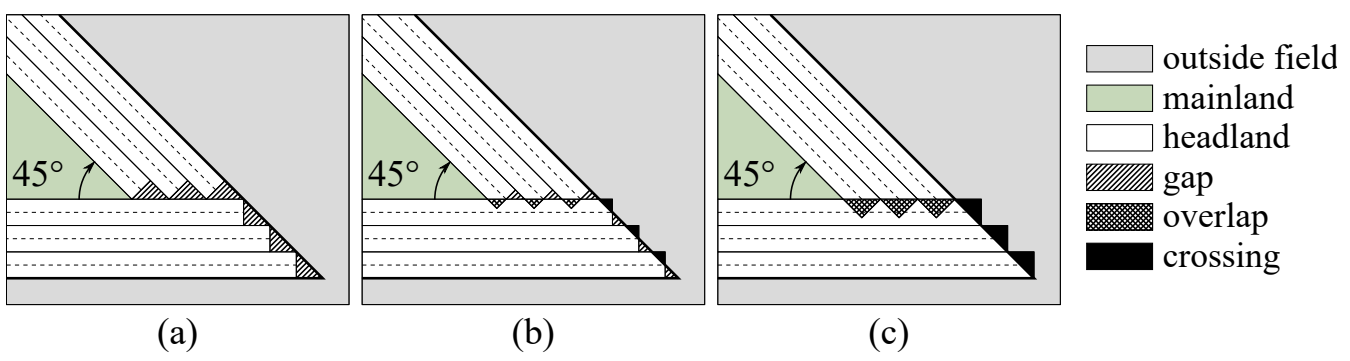

Fig. 2. Options to construct the tracks in field corners considering trade-off between gap, overlap, and crossing.

### 2.5 Modifying the headland track polygon corners

In the next and final step of the algorithm, the headland track polygons are cut and modified at all critical vertices to construct the final headland tracks. This is the key step of the algorithm to construct headland tracks that accommodate for turns with reversing, and the implement to reach the field corners.

There are three distinct ways to tweak the construction of the headland tracks in the field corners based on the desired trade-off between gap, overlap, and crossing the field boundary (illustrated in Fig. 2).

With tweak one, in a field corner such as shown in Fig. 2, all overlap and crossing can be avoided by constructing headland tracks that do not reach the field boundary (Fig. 2(a)). However, this results in gap in the field corners. With tweak two, when the headland tracks are extended exactly to the field boundary, the area of gap decreases (see Fig. 2(b)). However, this results in both overlap and crossing. With tweak three, by extending the headland tracks beyond the field boundary, all gap can be avoided (see Fig. 2(c)). However, the area of overlap and crossing increases.

In this work, the second tweak is implemented. With this tweak, in a simple convex field corner such as shown in Fig. 2, the combined area of gap, overlap, and crossing consist of 50% gap, 25% overlap, and 25% crossing.

The process of constructing the headland tracks according to the second tweak is illustrated in Fig. 3 that shows a field polygon (left) and a magnification of a convex corner of the same field polygon (right). Fig. 3(a) shows the critical vertices, Fig. 3(b) shows where the headland track polygons are cut and extended, and Fig. 3(c) shows the resulting headland tracks.

The headland is bounded by the inner and outer headland boundary. The outer headland boundary coincides with the field boundary, and the inner headland boundary coincides with the headland-mainland boundary. These boundaries are used to determine where the headland track polygons should be cut and extended.

First, the headland track polygons are cut at every critical vertex, resulting in as many independent polylines per headland track polygon as there are critical vertices in the polygon. The critical vertices are shown with red in Fig. 3(a). Each polyline is then cut again at the other end and extended at the other end, depending on the driving direction and whether the implement is mounted in the front or the rear of the vehicle. The example in Fig. 3 can be driven either clockwise with a rear mounted implement or counter-clockwise with a front mounted implement.

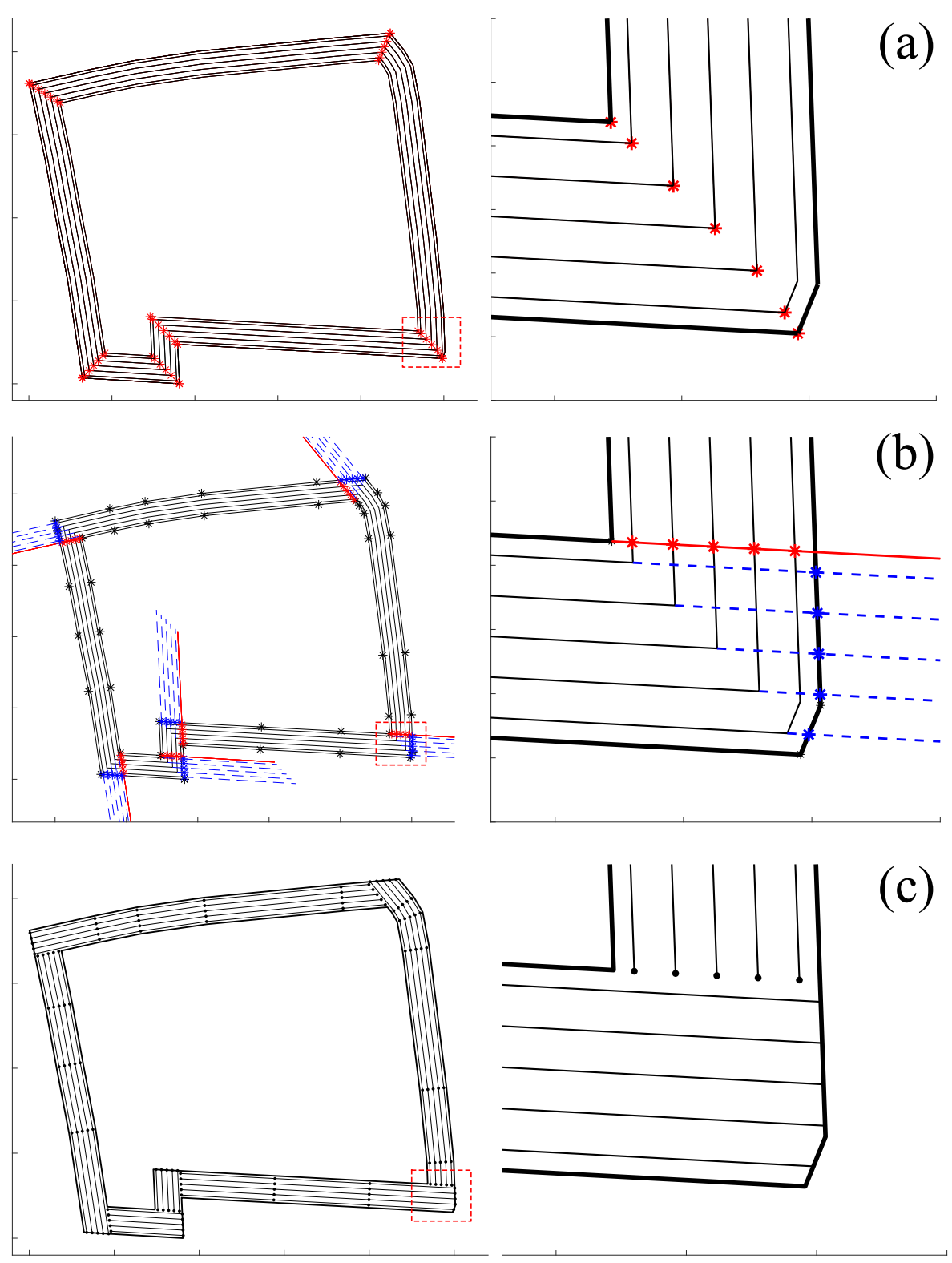

Fig. 3. Cutting and modifying the polygons: critical vertices (a), cutting and extending (b), end result (c).

Two polylines meet at every critical vertex. At a convex field corner, the first polyline is cut at the intersection with a line that extends from the inner headland boundary, shown with solid red line in Fig. 3(b). The line segment between the cut point and the critical vertex is discarded. The second polyline is extended from the critical vertex to the outer headland boundary, shown with a dashed blue line in Fig. 3(b). At a concave corner the roles of the inner and outer headland boundary are simply switched: cut at the outer boundary and extend to the inner boundary.

The left side of Fig. 3 shows the process and result for the entire field. A region around the bottom right corner of the field is highlighted with red dashed line. A magnification of this region is shown on the right side of Fig. 3.

One special case occurs when right after the critical vertex there are non-critical vertices that make the extend lines not parallel to the previous cut line. In these cases, the algorithm finds the next segment of the headland track polygon that is parallel to the previous cut line, starts the extension from the first vertex of this segment instead of the critical vertex, and makes the extension (dashed blue line) in the direction of the cut line (solid red line). Vertices in-between this vertex and the extend point are discarded. In these corners of the field, some additional gap, overlap, and crossing might occur.

## 3. ALGORITHM PROPERTIES

In addition to testing the proposed method on real field boundary data, the theoretical performance of the proposed method is compared to two other methods. For the sake of clarity and brevity, the following naming is used for the three methods in the rest of this paper. The approach where headland tracks are created as nested offset polygons and the corners are rounded with suitable curves, is referred to as “method A”. The approach adapted from the X-type turn by Mier et al. (2025) is referred to as “method B”. The proposed approach, where the nested offset polygon headland tracks are cut and modified so that the vehicle can reverse to the field corners with a suitable turn path planning method, is referred to as ”method C”.

The metrics used to assess the performance are the area of gap, overlap, and crossing the field boundary measured in square metres. Gap, overlap, and crossing are calculated based on geometry, i.e. the “expected result” of each method, assuming that the vehicle-implement combination is able to follow the path exactly as intended.

Gap means field area that should be worked on but was not. Gap results in a decreased yield, depending on the operation that was supposed to be done. Overlap means area that was worked on more than once. Overlap can be inefficient, and can occur both on the headland and the mainland. Crossing means crossing the field boundary, i.e. the implement has been outside the assigned workspace. Crossing is a safety concern especially if it is unknown what lies beyond the field boundary.

Fig. 4 illustrates the expected result of methods A, B, and C, respectively, at vertex angles of 45, 90 135, 225, 270, and 315 degrees for a predetermined implement operating width, vehicle-implement combination length, and minimum turning radius. The implement is mounted in the front of the vehicle and the vehicle is driving counter clockwise, approaching the field corner along the horizontal headland track.

The performance of each method, i.e. the resulting area of gap, overlap, and crossing, depends on the field corner vertex angle. To compare the three methods, the area of gap, overlap, and crossing is calculated for all methods, in field corners with one vertex such as in Fig. 4. The area of gap, overlap, and crossing is calculated for vertex angles between 30 and 330 degrees with 1 degree intervals.

The operating area of the implement is represented as a straight line along its operating width. The covered area is calculated based on the centre point of the line following the headland track while the line is orthogonal to the headland track. The covered area is the area that the line sweeps over while following each headland track. The intersection of the covered area with itself, the mainland, the headland, and the region outside the field boundary determines the area of overlap, gap, and crossing.

The calculation depends on the dimensions and minimum turning radius of the vehicle, the dimensions of the implement, and the number of headland tracks. In the example scenario a tractor with a 4-meter turning radius is mowing with a front mounted disc mower with an operating width of 3.16 meters. The tractor has a transport length of approximately 4.3 meters and a transport width of approximately 2.3 meters. Four headland tracks are generated, resulting in a headland width of 12.64 meters. When the mower is mounted in the front of the tractor the distance between the rear of the tractor and the centreline along the operating width of the mower is approximately 5.5 meters.

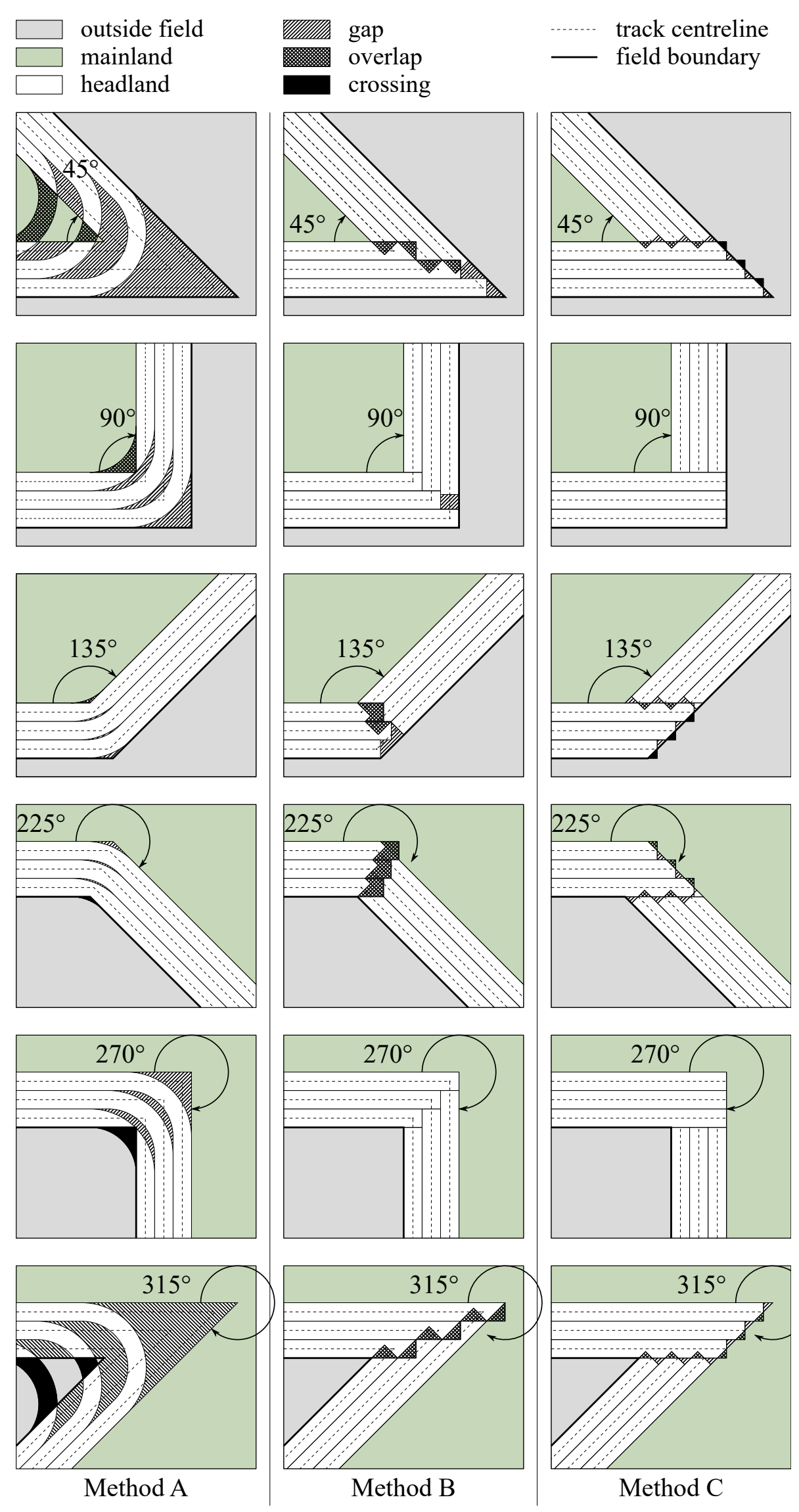

Fig. 4. Illustration of gap, overlap, and crossing in field corners with different vertex angles.

For method A, the vertices of the headland track polygons are assumed to be rounded with circular arcs. This assumes infinite steering speed (or stopping while reorienting the wheels), i.e. the vehicle is able to instantly transition from a straight path to a circular arc of maximum curvature, and the implement can operate during such a turn.

For method B, an adaptation of the X-type turn by Mier et al. (2025) is used. The original method requires an edge offset for the corner turns, and the gap that results from this offset depends on the length of the field sides. In an effort to make the methods meaningfully comparable, the edge offset has been omitted here. However, because of how the headland tracks are constructed in the field corners, not all areas of convex field corners can be covered due to the size of the vehicle (see Fig. 5).

For method C, the criticality of the vertices is not considered, i.e. the calculation is done as if the vehicle always makes a turn at the vertex independent of the vertex angle. This will show the performance of method C, independent of the choice of the critical angle threshold.

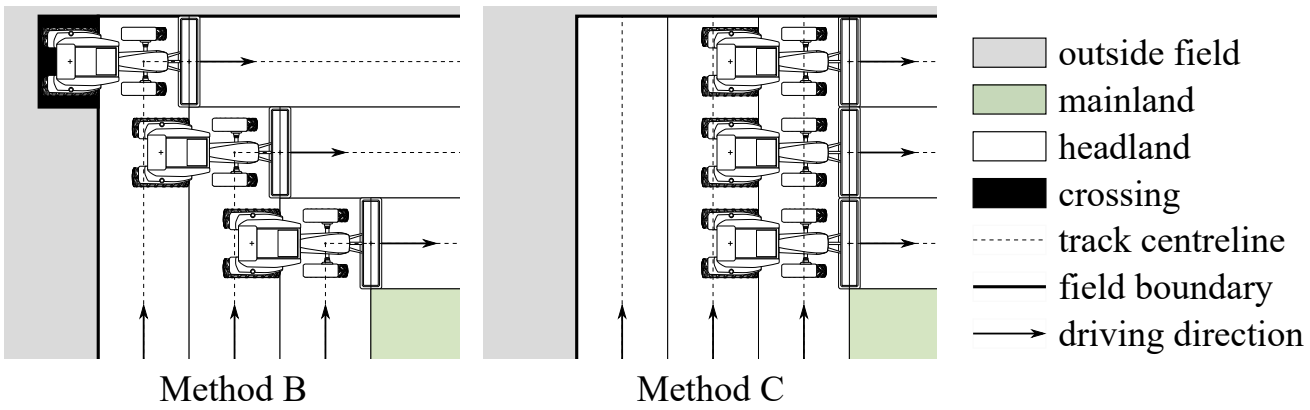

Fig. 5. Comparison of method B and method C headland tracks in field corners.

The result is shown in Fig. 6. With regards to gap, in field corners between 35 and 120 degrees method C outperforms both method A and method B. Between 30 and 35 degrees method B results in less gap than method A or method C. Between 120 and 180 degrees method A results in the smallest area of gap. Above 180 degrees method B leaves no gap, however, at the same time it results in more overlap than method A or method C. Below 120 degrees and above 240 degrees the gap that is observed with method A increases notably. In order for method B to cover the headland completely in concave field corners, the vehicle itself needs to cross the headland-mainland boundary.

By design, method B never crosses the field boundary, and therefore has zero crossing. With method C, the area of crossing goes to zero in concave field corners. With method B and method C, in field corners of 180 degrees and above, overlap occurs instead. With method A, crossing occurs in concave field corners, with its area increasing towards 330 degrees. Concerning crossing the field boundary, method B outperforms both method A and method C, at the cost of additional gap in convex field corners.

Method A results in crossing in concave field corners and overlap in convex field corners. While method A mostly outperforms method B and method C concerning overlap, it comes at the cost of more gap, especially in field corners below 120 degrees and above 240 degrees.

In field corners between 128 and 180 degrees, method A has the best performance concerning all three metrics (gap, overlap, crossing). Method C can provide most gains in corners close to 90 degrees and 270 degrees. While method B leaves no gap in concave field corners, otherwise method C performs better than method A or method B in field corners above 240 degrees.

This assessment applies as is for the example scenario. With a different implement operating width or different number of headland track polygons, the shape of the plots in Fig. 6 remains similar, but the exact comparison between the three methods might change.

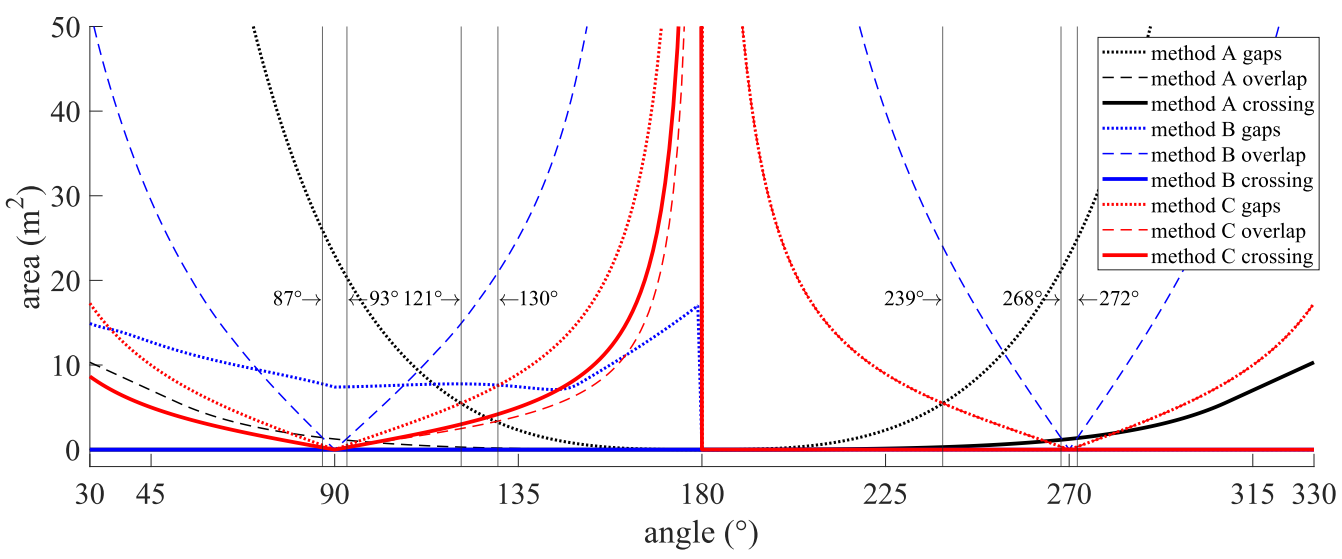

Fig. 6. Area of gap, overlap, and crossing in field corners between 30 to 330 degrees for all three methods.

## 4. EXPERIMENT

The proposed method (method C) is also tested on a set of real field data. The field shapes are from the dataset by Seyyedhasani et al. (2019) that contains 101 fields. The geographic coordinate data of the dataset have been converted to planar coordinates in the UTM system.

The tests are run in the same example scenario introduced before. The critical angle threshold is set to 40 degrees, which means that the tractor has to make a turn at any vertex with a vertex angle below 140 degrees or above 220 degrees. The parameter values for the example scenario are therefore $n_t = 4$, $w_{o,i} = 3.16m$, and $\gamma_t = 40\,\text{deg}$.

With these parameter values the following characteristics for the fields in the dataset are found (see table 1). Altogether 36 of the 101 fields have at least one of these occurrences. Therefore, 65 field shapes remain that the proposed method is expected to solve successfully, given the chosen parameter values. In the experiment, the implementation of the proposed method is tested on each of the 65 field shapes.

Table 1. Field shape characteristics given $n_t = 4$, $w_{o,i} = 3.16m$, and $\gamma_t = 40\,\text{deg}$.

| Characteristic | Number of fields |
|---|---|
| Critical vertex appears | 16 |
| Critical vertex disappears | 23 |
| Narrow region | 9 |
| Any of the above three occurs | 36 |

## 5. RESULTS

The implementation of the proposed method is able to solve most of the 65 field shapes. Three examples of successful solutions are visualised in Fig. 7. A thicker line shows the field boundary and the headland-mainland boundary, while the headland tracks between the boundaries are drawn with a thinner line. The dots show individual vertices on the headland tracks.

Fig. 7 (a) shows field corners with multiple vertices. Only one of the vertices is a critical vertex, and the implementation successfully extends the headland tracks to the field boundary considering the other vertices. Fig. 7 (b) shows regions with multiple nearby field corners where the implementation can solve the headland tracks successfully. Fig. 7 (c) shows how the implementation solves headland tracks with curved contours that consist of several vertices.

The implementation was not able to solve five of the 65 fields, due to several nearby alternating convex and concave critical vertices. These critical vertices are so near to each other that the line to extend some of the headland tracks will not encounter the outer or inner headland boundary before intersecting other headland tracks.

## 6. DISCUSSION

Based on experiments with several different field boundary polygons, the proposed method can successfully plan coverage paths for headlands. In comparison to two other methods, adapted from earlier literature, it provided a better coverage result especially in field corners of around 90 degrees, and 240 degrees and above, measured by the area of gap, overlap, and crossing the field boundary. However, the exact numbers depend on the specific scenario.

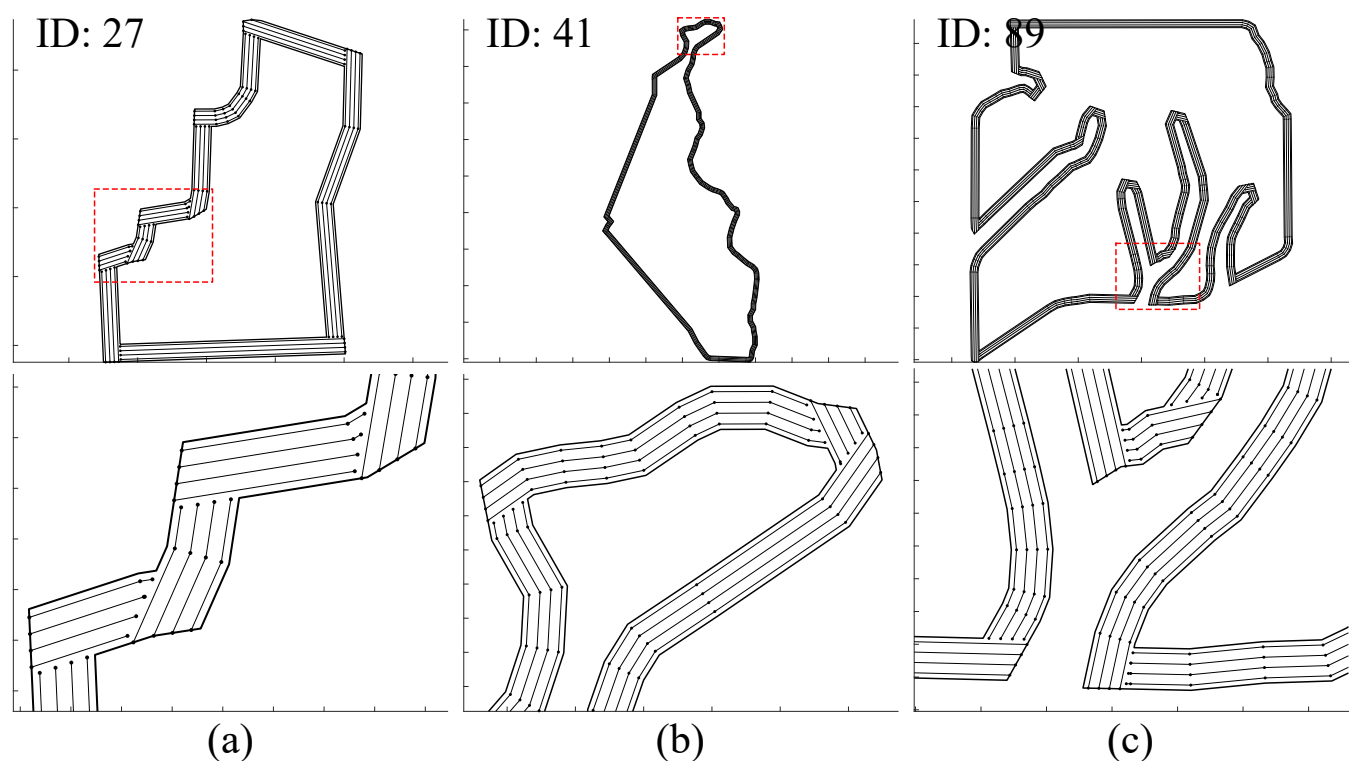

Fig. 7. Solutions by the proposed method. A magnification of the region enclosed by red dashed line below.

The ability to solve the headland coverage was found to be affiliated with specific characteristics of the field boundary, and all methods were found to have their advantages and disadvantages. Some special cases remain to be solved based on the comparison and experiments.

One challenge occurs with nearby critical vetices, when very short path segments are needed due to the short distances. Therefore, the sides of the field are assumed to be long enough. A similar assumption is made by Mier et al. (2025). A second, related challenge occurs when the critical vertices alternate between convex and concave. In particular, the proposed method might fail to extend the headland tracks in this case, due to the extended headland track intersecting another headland track before reaching the headland boundary. A third challenge occurs with narrow passages that do not fit multiple headland tracks. This means that the required amount of headland tracks cannot be constructed in the narrow region. Finally, a fourth challenge occurs while creating the nested headland track polygons, when critical vertices appear or disappear. When a critical vertex appears or disappears, the requirement for a turn in the corresponding corner of the headland track polygon changes as a result. None of the examined methods can solve some of these cases, potential solutions to which are left for future work.

Finally, the proposed method considers two cases that the other methods do not explicitly take into account. First, it is able to solve field corners with multiple vertices, assuming that only one of them is critical. It also allows transitioning between the inner and outer headland tracks by design, since the transition can take place in any field corner where a turn is needed anyway.

Following these insights, we suggest a mix of the three methods to create a full feasible coverage path plan for the headland. Based on a case-by-case analysis of the gap, overlap, crossing, and the physical limits of the vehicle-implement combination, the critical angle threshold can be adapted for this use, to apply each method according to where they provide the better trade-off.

## REFERENCES

Backman, J., Piirainen, P., and Oksanen, T. (2015). Smooth turning path generation for agricultural vehicles in headlands. *Biosyst. Eng.*, 139, 76–86. doi:10.1016/j.biosystemseng.2015.08.005.

Bochtis, D. and Vougioukas, S. (2008). Minimising the non-working distance travelled by machines operating in a headland field pattern. *Biosyst. Eng.*, 101, 1–12. doi:10.1016/j.biosystemseng.2008.06.008.

Edwards, G.T., Hinge, J., Skou-Nielsen, N., Villa-Henriksen, A., Sørensen, C.A.G., and Green, O. (2017). Route planning evaluation of a prototype optimised infield route planner for neutral material flow agricultural operations. *Biosyst. Eng.*, 153, 149–157. doi:10.1016/j.biosystemseng.2016.10.007.

Hameed, I., Bochtis, D., Sørensen, C., and Nøremark, M. (2010). Automated generation of guidance lines for operational field planning. *Biosyst. Eng.*, 107(4), 294–306. doi:10.1016/j.biosystemseng.2010.09.001.

Höffmann, M., Patel, S., and Büskens, C. (2022). Weight-optimized nurbs curves: Headland paths for nonholonomic field robots. In *ICARA 2022*, 81–85. doi:10.1109/ICARA55094.2022.9738525.

Höffmann, M., Patel, S., and Büskens, C. (2024). Optimal guidance track generation for precision agriculture: A review of coverage path planning techniques. *J. Field Robot.*, 41(3), 823–844. doi:10.1002/rob.22286.

Jeon, C.W., Kim, H.J., Yun, C., Han, X., and Kim, J.H. (2021). Design and validation testing of a complete paddy field-coverage path planner for a fully autonomous tillage tractor. *Biosyst. Eng.*, 208, 79–97. doi:10.1016/j.biosystemseng.2021.05.008.

Mier, G., Fennema, R., Valente, J., and de Bruin, S. (2025). Continuous curvature path planning for headland coverage with agricultural robots. *J. Field Robot.* doi:10.1002/rob.22489.

Nilsson, R.S. and Zhou, K. (2020). Method and benchmarking framework for coverage path planning in arable farming. *Biosyst. Eng.*, 198, 248–265. doi:10.1016/j.biosystemseng.2020.08.007.

Pour Arab, D., Spisser, M., and Essert, C. (2023). Complete coverage path planning for wheeled agricultural robots. *J. Field Robot.*, 40(6), 1460–1503. doi:10.1002/rob.22187.

Seyyedhasani, H., Dvorak, J.S., and Roemmele, E. (2019). Routing algorithm selection for field coverage planning based on field shape and fleet size. *Comput. Electron. Agric.*, 156, 523–529. doi:10.1016/j.compag.2018.12.002.

Soitinaho, R., Väyrynen, V., and Oksanen, T. (2024). Heuristic cooperative coverage path planning for multiple autonomous agricultural field machines performing sequentially dependent tasks of different working widths and turn characteristics. *Biosyst. Eng.*, 242, 16–28. doi:10.1016/j.biosystemseng.2024.04.007.

Väyrynen, V.T. (2019). Feasible real-time turn path generation for agricultural field machines with spatial constraints. Master's thesis, Aalto University, Helsinki.

Zhou, K., Jensen, A.L., Sørensen, C., Busato, B., and Bochtis, D. (2014). Agricultural operations planning in fields with multiple obstacle areas. *Comput. Electron. Agric.*, 109, 12–22. doi:10.1016/j.compag.2014.08.013.